\documentclass[twoside,leqno,twocolumn]{article}

\usepackage[letterpaper]{geometry}

\usepackage{ltexpprt}
\usepackage{hyperref}
\usepackage{subfigure}
\usepackage{adjustbox}
\usepackage{cite}
\usepackage{amsmath,amssymb,amsfonts}
\usepackage{graphicx}
\usepackage{textcomp}
\usepackage{xcolor}
\usepackage{xspace}
\usepackage{graphicx}
\usepackage{mathtools}
\usepackage{amsfonts}
\usepackage{amsmath}
\usepackage{amssymb}
\usepackage{algorithm}
\usepackage{algpseudocode}

\usepackage{makecell}
\usepackage{colortbl}
\usepackage{amssymb}
\usepackage{balance}

\newcommand*\Input[1]{\Statex \textbf{Input:} #1}
\newcommand*\Output[1]{\Statex \textbf{Output:} #1}
\algrenewcommand\alglinenumber[1]{#1}

\usepackage{amsthm}
\usepackage{graphicx}
\usepackage{multirow}
\usepackage{tabularx}
\usepackage{booktabs}

\newtheoremstyle{def_style}
  {}          
  {}          
  {}          
  {}          
  {\bfseries} 
  {.}         
  {.5em}      
  {}          

\theoremstyle{def_style}

\theoremstyle{def_style}

\theoremstyle{def_style}

\theoremstyle{def_style}

\theoremstyle{def_style}

\DeclareMathOperator*{\argmax}{arg\,max}

\newcommand{\model}{\textsc{AutoFM}\xspace}

\begin{document}

\title{Automated Fusion of Multimodal Electronic Health Records for Better Medical Predictions}
\date{}

\author{Suhan Cui$^{1}$, Jiaqi Wang$^1$, Yuan Zhong$^1$, Han Liu$^2$, Ting Wang$^3$, Fenglong Ma$^1$\thanks{Corresponding author.}\\
\small\baselineskip=9pt $^1$Pennsylvania State University,
\small\baselineskip=9pt $^2$Dalian University of Technology, $^3$Stony Brook University\\
\small  $^1$\{suhan, jqwang, yfz5556, fenglong\}@psu.edu, $^2$liu.han.dut@gmail.com, $^3$twang@cs.stonybrook.edu
}

\maketitle

\fancyfoot[R]{\scriptsize{Copyright \textcopyright\ 2024 by SIAM\\
Unauthorized reproduction of this article is prohibited}}

\begin{abstract} 
The widespread adoption of Electronic Health Record (EHR) systems in healthcare institutes has generated vast amounts of medical data, offering significant opportunities for improving healthcare services through deep learning techniques. However, the complex and diverse modalities and feature structures in real-world EHR data pose great challenges for deep learning model design. To address the multi-modality challenge in EHR data, current approaches primarily rely on hand-crafted model architectures based on intuition and empirical experiences, leading to sub-optimal model architectures and limited performance. Therefore, to automate the process of model design for mining EHR data, we propose a novel neural architecture search (NAS) framework named \model, which can automatically search for the optimal model architectures for encoding diverse input modalities and fusion strategies. We conduct thorough experiments on real-world multi-modal EHR data and prediction tasks, and the results demonstrate that our framework not only achieves significant performance improvement over existing state-of-the-art methods but also discovers meaningful network architectures effectively\footnote{Source code can be found via the link: \url{https://github.com/SH-Src/AUTOMF}}.
\end{abstract}

\noindent\textbf{\small Keywords:} Electronic Health Records, Neural Architecture Search, Multimodal Fusion

\section{Introduction}


Electronic Health Record (EHR) systems have been extensively adopted in numerous hospitals and healthcare institutions, resulting in the generation of vast amounts of patient EHR data on a daily basis. This data holds significant potential for various predictive tasks, including but not limited to diagnosis prediction~\cite{choi2016doctor, ma2017dipole}, medical recommendation \cite{shang2019gamenet}, health risk prediction \cite{choi2016retain,ma2018risk}, and hospital readmission \cite{huang2019clinicalbert}. However, the heterogeneous and multimodal nature of EHR data poses significant challenges in the design of effective deep predictive models.



Most existing studies~\cite{xu2018raim, feng2019dcmn, qiao2019mnn, yang2021leverage} primarily focus on the design of \emph{hand-crafted} model architectures for integrating multimodal EHR data. However, this approach necessitates considerable domain expertise, which may introduce the potential for human bias. To address these concerns, recent research~\cite{xu2021mufasa, cui2022automed} has proposed the utilization of neural architecture search (NAS) techniques~\cite{elsken2019neural} to automatically search for suitable architectures for modeling multimodal EHR data, thereby eliminate the need for human intervention. Although these methods demonstrate improved performance compared to hand-crafted approaches, they still encounter certain issues:

\paragraph{C1 -- Diversifying the Search Space}
Current approaches~\cite{xu2021mufasa, cui2022automed} typically employ a uniform search space across all modalities when conducting neural architecture search (NAS). However, EHR data encompasses diverse modalities, including tabular demographics, discrete medical codes, continuous monitoring data from ICU stays, and unstructured clinical notes. A uniform search space fails to adequately capture the heterogeneity of these modalities.
Furthermore, existing methods primarily utilize simple operations like concatenation and addition within the fusion search space, which may not effectively capture the complex interactions among different EHR modalities. Therefore, it is crucial to explore more suitable search spaces for multimodal feature encoding and fusion that can better accommodate the diverse nature of EHR modalities.


\paragraph{C2 -- Customizing the Search Optimization}
Existing approaches utilize either evolutionary NAS~\cite{real2019regularized} or differentiable architecture search (DARTS)~\cite{liu2018darts} to find the optimal architecture within the defined search space. Although DARTS represents a significant efficiency improvement compared to evolutionary NAS, it often encounters issues of robustness~\cite{zela2019understanding} and may struggle to identify suitable architectures. Given the complexity of the search space in our specific task, directly applying DARTS may result in poor performance for the searched architectures. Consequently, a customized search algorithm is necessary to discover meaningful architectures for effectively fusing multimodal EHR data.


\paragraph{C3 -- Deriving the Optimal Architecture}
DARTS-based methods, such as AutoMed~\cite{cui2022automed}, determine the final architecture by selecting operations based on the magnitudes of architecture weights obtained from the trained supernet. However, it has been observed that the operation with the highest architectural weight on the supernet does not necessarily correspond to its actual contribution to the generalization performance~\cite{wang2021rethinking}. This arbitrary discretization approach can result in a significant drop in performance, a concern that becomes more pronounced when modeling multimodal EHR data due to the complexity of the search space. Thus, the development of a new discretization technique tailored to our specific task is necessary to derive an optimal architecture.



\paragraph{Our Approach}
To overcome the aforementioned challenges, we propose a novel neural architecture search (NAS) framework called \model for \underline{\textbf{Auto}}matically \underline{\textbf{F}}using \underline{\textbf{M}}ulti-modal EHR data. 
\model addresses \textbf{C1} by introducing a new two-stage search space. The first stage, called \emph{modality-specific search}, focuses on designing specialized encoding modules for each input modality, incorporating feature encoding and feature interaction operations to explore potential early fusion strategies. The second stage, known as \emph{multi-modal fusion search}, focuses on creating late fusion architectures. This involves a feature selector component to determine the selected modalities and a searchable fusion component to identify the optimal fusion operation.

To tackle \textbf{C2}, we introduce a customized loss term within the original bi-level optimization formulation of DARTS. This customized loss promotes diversity within the feature selector, guiding the search algorithm toward discovering more meaningful architectures.
To address \textbf{C3}, we devise a novel pruning-based algorithm to select the optimal architectures after training the supernet. This algorithm effectively preserves the performance of the derived architecture during the discretization process.
We extensively evaluate our proposed method on real-world multi-modal EHR data, showing its superiority over existing state-of-the-art models. Through these experiments, we demonstrate the effectiveness and advantages of our approach.

\begin{figure*}
    \centering
    \includegraphics[width=0.95\textwidth]{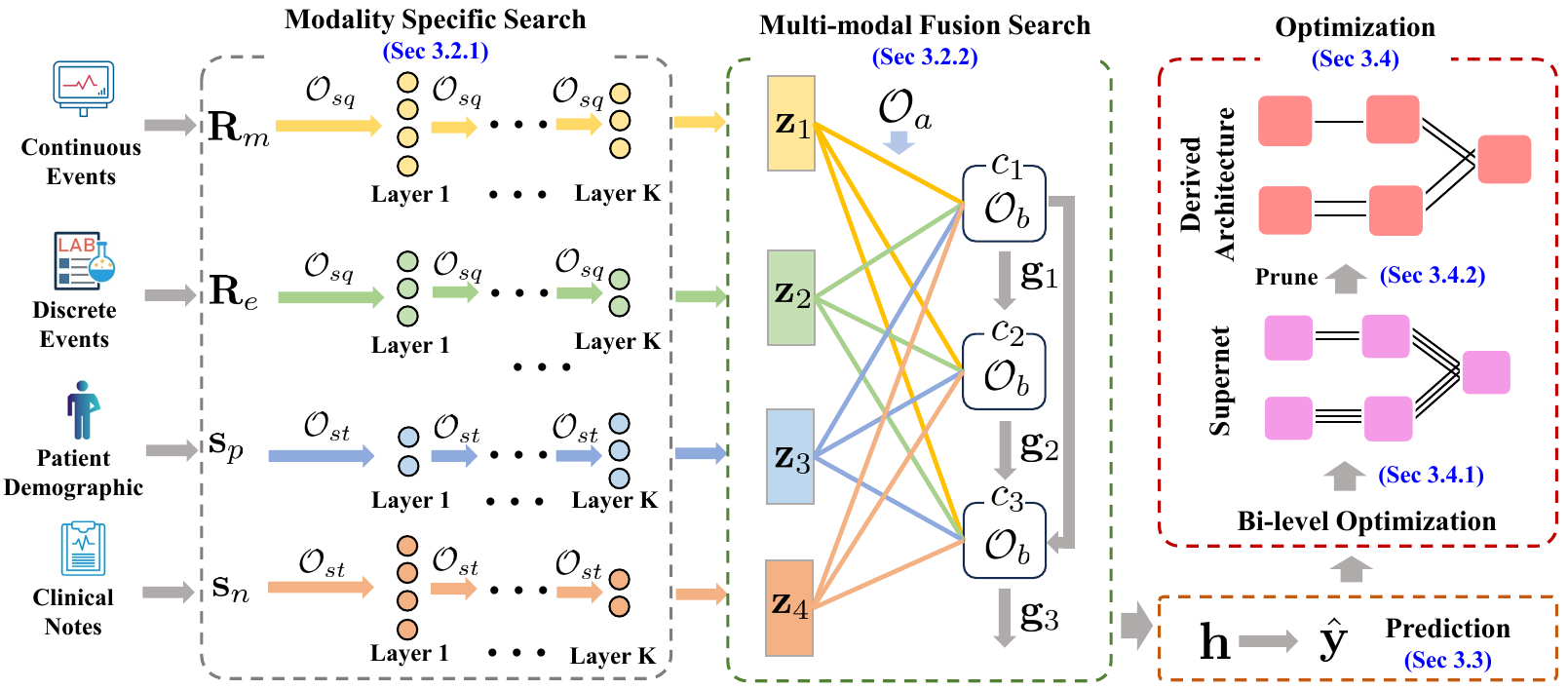}
    \vspace{-0.15in}
    \caption{Overview of the proposed \model.}
    \label{fig:framework}
    \vspace{-0.1in}
\end{figure*}
\section{Related Work}

\subsection{Modeling Multi-modal EHR data}

Several approaches have been proposed to leverage multi-modal electronic health record (EHR) data for enhanced predictive modeling~\cite{yang2021leverage, zhang2022improving}. For instance, studies such as RAIM \cite{xu2018raim}, DCMN \cite{feng2019dcmn}, and MNN \cite{qiao2019mnn} integrate various data sources, including patient monitoring data and clinical events, to predict outcomes such as ICU stay length, mortality, and diagnoses. However, these approaches rely on hand-crafted neural networks, limiting their applicability to pre-defined modalities and tasks.
In contrast, MUFASA~\cite{xu2021mufasa} and AutoMed~\cite{cui2022automed}  employ neural architecture search (NAS) methods to discover more suitable model architectures for multi-modal EHR data. However, these methods either come with high computational costs or impose constraints on input features and fusion strategies.

Our work distinguishes itself by introducing a comprehensive search space encompassing modality-specific architectures and diverse multi-modal fusion strategies. Besides, we propose a new optimization loss on top of the more efficient DARTS \cite{liu2018darts} algorithm, enabling the discovery of more reasonable and advantageous architectures for multi-modal EHR data.

\subsection{Neural Architecture Search}

Neural architecture search (NAS) \cite{elsken2019neural} automates the process of finding optimal deep neural network architectures. Different approaches such as reinforcement learning \cite{zoph2016neural} and evolutionary search \cite{real2019regularized} have been used, but they require significant computational costs. To improve efficiency, more advanced techniques have been employed, such as weight sharing \cite{pham2018efficient}, sequential model-based optimization \cite{liu2018progressive}, and Bayesian optimization \cite{kandasamy2018neural}. A recent approach called differentiable architecture search (DARTS) \cite{liu2018darts} has emerged, which achieves high search efficiency by using a continuous relaxation of the discrete architecture. 
This work addresses the limitations of DARTS when applied to multi-modal EHR data. It introduces an additional penalty during supernet training to encourage the selection of more meaningful architectures. Additionally, a new pruning-based architecture selection method is proposed to maintain the supernet's performance during discretization.

\section{Methodology}

As depicted in Figure~\ref{fig:framework}, our proposed {\model} framework takes multiple heterogeneous EHR data as input. To effectively handle this diverse data, we introduce a novel multimodal search space that enables the automatic search for optimal architectures for modality representation and fusion strategies. This search space allows our framework to discover the most suitable approaches for representing each modality and effectively fuse them together for improved predictive modeling.

\subsection{Multimodal EHR Data Embedding}
Our model takes four modalities as the input, including two types of \textbf{sequential} data -- (1) continuous events $\mathbf{M} = [\mathbf{m}_1, \cdots, \mathbf{m}_T] \in \mathbb{R}^{d_1 \times T}$ and (2) discrete events $\mathbf{E}=[\mathbf{e}_1, \cdots, \mathbf{e}_T] \in \mathbb{R}^{d_2 \times T}$ and two types of \textbf{static} data -- (3) patient demographics $\mathbf{p} \in \mathbb{R}^{d_3}$ and (4) the corresponding clinical note $\mathbf{n} \in \mathbb{R}^{d_4}$, where $T$ is the number of time slots and $\mathbf{n}$ is the output of \emph{[CLS]} from the pre-trained language model - ClinicalBERT \cite{huang2019clinicalbert}. The details of data processing can be found in Section~\ref{sec:data}.

We first map these features into the same latent space by a fully connected layer:
\begin{align}
\small
    \mathbf{R}_m &= \mathbf{W}_m^\top[\mathbf{m}_1, \cdots, \mathbf{m}_T] + \mathbf{b}_m,\\
    \mathbf{R}_e &= \mathbf{W}_e^\top[\mathbf{e}_1, \cdots, \mathbf{e}_T]  + \mathbf{b}_e,\\
    \mathbf{s}_p &= \mathbf{W}_p^\top \mathbf{p} + \mathbf{b}_p,\\
    \mathbf{s}_n &= \mathbf{W}_n^\top \mathbf{n} + \mathbf{b}_n,
\end{align}
where $\mathbf{W}_m \in \mathbb{R}^{d_1 \times d_e}$, $\mathbf{W}_e \in \mathbb{R}^{d_2 \times d_e}$, $\mathbf{W}_p \in \mathbb{R}^{d_3 \times d_e}$, $\mathbf{W}_n \in \mathbb{R}^{d_4 \times d_e}$, $\mathbf{b}_e \in \mathbb{R}^{d_e}$, $\mathbf{b}_m \in \mathbb{R}^{d_e}$, $\mathbf{b}_p \in \mathbb{R}^{d_e}$, and $\mathbf{b}_n \in \mathbb{R}^{d_e}$ are learnable parameters.

\begin{table*}[t]
    \centering
    \caption{Modality-specific operations.}
    \label{tab:modality_search_operations}
    \resizebox{1\textwidth}{!}{
    \begin{tabular}{c|l|l}
    \hline
         \textbf{Feature} &  \textbf{Operation} & \textbf{Operation Description}\\\hline\hline
         \multirow{10}{*}{\makecell[c]{Static\\($\mathcal{O}_{st}$)}}
         & Identity & This is a default operation that does not transform the input feature.\\\cline{2-3}
         &Linear Layer & \makecell[l]{Apply linear transformation with ReLU activation, where $\mathbf{W}_1\in\mathbb{R}^{d_e\times d_e}$ and $\mathbf{b}_1\in\mathbb{R}^{d_e}$:\\ $o(x) = {\rm ReLU} (\mathbf{W}_1 x + \mathbf{b}_1)$}\\ \cline{2-3}
         & \makecell[l]{Static-Static\\Feature Interaction} & \makecell[l]{Apply concatenation and feed-forward layer to model the interaction between the current \\feature and the other static feature, where $\mathbf{W}_2\in\mathbb{R}^{2d_e\times d_e}$, $\mathbf{b}_2\in\mathbb{R}^{d_e}$ and $x^\prime$ is the input feature \\of the other static modality $\mathbf{s}_p$ or $\mathbf{s}_n$: $o(x) = \mathbf{W}_2[x;x^\prime] + \mathbf{b}_2$}\\ \cline{2-3}
         & \makecell[l]{Static-Sequential\\Feature Interaction} & \makecell[l]{Apply the attention mechanism to model the interaction with the current feature as the query \\and the other features as the keys. Since we have two sequence features, we then have two \\interaction operations, where $\mathbf{W}_{q}, \mathbf{W}_{k}, \mathbf{W}_{v}\in\mathbb{R}^{d_e\times d_e}$: \\$o(x) = \;{\rm Softmax}\left(\frac{(\mathbf{W}_{q}x)^\top\cdot(\mathbf{W}_{k}x^\prime)}{\sqrt{d_e}}\right)\cdot
                  (\mathbf{W}_{v}x^\prime), x^\prime \in \{\mathbf{R}_m, \mathbf{R}_e\}$}\\ \hline\hline

         \multirow{13}{*}{\makecell[c]{Sequential\\($\mathcal{O}_{sq}$)}}
         & Identity & This is a default operation that does not transform the input feature.\\\cline{2-3}
         & Recurrent Layer & \makecell[l]{Apply Recurrent Neural Network(RNN) to process the input feature, which is able to capture \\the temporal patterns. In our model, we use Gated Recurrent Unit (GRU) \cite{chung2014empirical} \\as the basic operation.}\\ \cline{2-3}
         & Self-Attention Layer & \makecell[l]{Apply self-attention mechanism \cite{vaswani2017attention} to model the long-term dependencies \\of the sequence feature. }\\ \cline{2-3}
         & 1-D Convolution & \makecell[l]{Apply 1-D convolution operation to process the feature and capture the local correlations. We \\use the padding and strides that can maintain the same shape as the input tensor.}\\ \cline{2-3}
         & Feed-forward Layer & Apply a linear layer to all positions of the sequence feature.\\ \cline{2-3}
         & \makecell[l]{Sequential-Sequential\\Feature Interaction} & \makecell[l]{Apply cross-modal attention mechanism to model the interaction between two sequence \\features. The current feature will serve as the queries, and the other feature will serve as the keys. \\The formulation is the same as Static-Sequential Feature Interaction except that the queries \\become sequential features this time. } \\
         \hline
    \end{tabular}
    }
    \vspace{-0.15in}
    
\end{table*}

\subsection{Multi-Modal Search Space Design}
Our approach incorporates two stages of searchable modules: (1) \textbf{modality-specific search} and (2) \textbf{ multimodal fusion search}. In the first stage, we focus on designing modality-specific search spaces tailored to each individual modality. These search spaces take into account not only the encoding of the current feature within a modality but also the potential early interactions between different modalities. This allows for the exploration of various encoding strategies and fusion techniques at the modality level.

Moving to the second stage, our approach encompasses a search space specifically designed for late fusion architectures. This search space consists of two crucial modules: the \emph{feature selector} and the \emph{searchable fusion module}. The feature selector component determines which modalities should be selected and incorporated into the fusion process. The searchable fusion module then identifies the optimal fusion operation to be applied. By jointly optimizing the feature selection and fusion operation within this search space, we enable the framework to automatically discover effective strategies for multi-modal feature fusion.

In the following sections, we will provide more detailed explanations of each stage, highlighting their respective contributions and functionalities.

\subsubsection{Modality Specific Search}

To facilitate the search process using differentiable methods, we assume that the architecture for each modality type consists of a network comprising $K$ sequentially connected layers, as illustrated in Figure \ref{fig:framework}. Consequently, our objective is to search for the optimal operation for each layer $k$. To make the search space compatible with differentiable search methods, we define a mixed operation at each layer. This mixed operation is a weighted sum of all operations within a predefined set $\mathcal{O}$. The formulation can be expressed as follows:
\begin{equation} \label{eq:mixed}
\begin{aligned}
    x_k &= \sum_{o\in \mathcal{O}}\frac{{\rm exp}(\alpha_{o}^{(k)})}{\sum_{o^\prime \in \mathcal{O}}{\rm exp}(\alpha_{o^\prime}^{(k)})}o(x_{k-1}), \\
    k &=1, \cdots, K,
\end{aligned}
\end{equation}
where $x_k$ represents the feature of layer $k$. $x_0$ can be any of the input features $[\mathbf{R}_m, \mathbf{R}_e, \mathbf{s}_p, \mathbf{s}_n]$. $\alpha$'s are architecture parameters that measure the weights for candidate operations, and we have four groups of $\alpha$'s in total for different modalities. 


Next, we list the candidate operation sets for each modality in Table~\ref{tab:modality_search_operations}. As we have two types of input modalities, we design two distinct modality-specific search spaces to cater to their specific characteristics. For both types of features, in addition to the feature encoding operations, we also incorporate \textbf{interaction operations}. These interaction operations handle the interactions between the current modality and other modalities, both within the same category and across different categories. By including these interaction operations, our method is capable of exploring how to effectively process all the input features and discover fine-grained early fusion operations among different modalities. This allows for comprehensive modeling and integration of the various modalities present in the EHR data.

\begin{table*}[t]
\centering
\caption{Fusion operations.}\label{tab:fusion_operation}

\resizebox{0.9\textwidth}{!}{
\begin{tabular}{l|l}
\hline
\textbf{Operation}       & \textbf{Operation Description} \\ \hline
Sum & \makecell[l]{This operation sums all the input features together to fuse them: $o(u_c) = \mathbf{u}_1^{(c)} + \mathbf{u}_2^{(c)} + \cdots + \mathbf{u}_{4 + c - 1}^{(c)}$} \\ \cline{1-2}
MLP        &  \makecell[l]{This operation applies a multi-layer perceptron to the sum of all input features, where $\mathbf{W}_3\in\mathbb{R}^{d_e\times d_e}$ \\ and $\mathbf{b}_3\in\mathbb{R}^{d_e}$: $o(u_c) = {\rm ReLU} (\mathbf{W}_3\mathbf{u}^{\prime(c)} + \mathbf{b}_3)$, and $\mathbf{u}^{\prime^(c)} = \mathbf{u}_1^{(c)} + \mathbf{u}_2^{(c)} + \cdots + \mathbf{u}_{4 + c - 1}^{(c)}$}                     \\ \cline{1-2}
\makecell[l]{Attentive \\Sum}&   \makecell[l]{This operation first uses linear projection to generate weights for all features and then applies \\a weighted sum to aggregate all features, where $\mathbf{W}_{\phi} \in \mathbb{R}^{d_e}$ and $b_{\phi} \in \mathbb{R}$: $o(u_c) = \sum_{i=1}^{4+c-1}\phi_i\mathbf{u}_{i}^{(c)},$ \\ $[\phi_1, \cdots, \phi_{4+c-1}] = {\rm Softmax}(l_1, \cdots, l_{4+c-1})$, and $l_i = \mathbf{W}_{\phi}\mathbf{u}_i^{(c)} + {b}_{\phi}, i = 1, \cdots, 4+c-1$}                    \\    \hline          
\end{tabular}
}

\end{table*}

\subsubsection{Multimodal Fusion Search}
Once we obtain the output features for all modalities from the first stage, we apply a fixed max pooling operation to the encodings of sequence features over the sequence length dimension. This step ensures that we have encodings of the same shape for all modalities, denoted as $[\mathbf{z}_1, \mathbf{z}_2, \mathbf{z}_3, \mathbf{z}_4]\in\mathbb{R}^{d_e}$. Next, we will discuss how to effectively fuse these features using searchable modules.

Our fusion module follows a directed acyclic graph (DAG) design. Given the input features $[\mathbf{z}_1, \cdots, \mathbf{z}_4]$, we add one computation node at each step, and the newly added nodes can connect to all previous nodes and features, as shown in Figure \ref{fig:framework}.

Assuming we have a total of $C$ nodes, at node $c$, we have the following features as inputs: $[\mathbf{z}_1, \cdots, \mathbf{z}_4, \mathbf{g}_1, \cdots, \mathbf{g}_{c-1}]$, where the $\mathbf{g}$'s represent the output features for the corresponding nodes. Our goal is to search for two operations: (1) selecting the previous features that should be taken as inputs to the current node, and (2) determining the fusion operation for the selected inputs. To accomplish these steps, we design two searchable modules: the feature selector and the searchable fusion module. These modules enable us to dynamically and adaptively determine the most relevant features and fusion operations at each node of the fusion module.

\paragraph{Feature Selector} 
To handle the feature selection at each node $c$, we define an operation set $\mathcal{O}_a$ that consists of only two operations: \textbf{Identity} and \textbf{Zero}. These operations determine whether to select the corresponding feature or not. Similar to Eq.~\eqref{eq:mixed}, we apply the mixed operation over $\mathcal{O}_a$ to each input feature for node $c$. As a result, at node $c$, we obtain a list of $(4 + c - 1)$ features, denoted as $u_c = [\mathbf{u}_1^{(c)}, \cdots, \mathbf{u}_{4 + c - 1}^{(c)}]$, where the values of these features can be either 0 or the original values. This process allows us to dynamically select relevant features based on the search algorithm's decisions at each node.

\paragraph{Searchable Fusion} 
To determine the fusion operation for the selected features at each node $c$, we define a candidate set $\mathcal{O}_b$ that contains different fusion strategies, as listed in Table~\ref{tab:fusion_operation}. These fusion strategies represent different ways of combining the selected features. Similar to the feature selection step, we apply mixed operations over $\mathcal{O}_b$ at each node to determine the fusion operation. This process allows us to explore and search for the optimal fusion strategy that effectively combines the selected features.


\subsection{Prediction}
To obtain a comprehensive representation of the entire EHR data, we linearly combine all the node features from the multi-modal fusion module $[\mathbf{g}_1, \cdots, \mathbf{g}_C]$, i.e., $\mathbf{h} = \sum_{c=1}^{C}\mathbf{w}_{c}\mathbf{g}_{c}$, where $\mathbf{w}_{c}\in\mathbb{R}$ is the learned weight.


For the binary classification problem, we use sigmoid as the activation function to calculate the prediction probability as follows:
\begin{equation}
    \hat{{y}} = {\rm sigmoid}(\mathbf{W}_y\mathbf{h} + \mathbf{b}_y).
\end{equation}
For the multi-label classification problem, we use softmax to generate the probability score for all classes as follow: 
\begin{equation}
    \hat{\mathbf{y}} = {\rm softmax}(\mathbf{W}_y\mathbf{h} + \mathbf{b}_y),
\end{equation}
where $\hat{\mathbf{y}}\in\mathbb{R}^{P}$ and $P$ is the number of classes.

\subsection{Optimization} \label{sec:optimize}

\subsubsection{Supernet Training} \label{sec:supernet}
We use the bi-level optimization technique as DARTS~\cite{liu2018darts} to optimize the model architecture weights and all other learnable parameters simultaneously:
\begin{equation} \label{eq:bilevel}\small
\begin{aligned}
&\min_{\boldsymbol{\alpha}}\mathcal{L}_{val}(\mathbf{W}^*(\boldsymbol{\alpha}, \boldsymbol{\beta}, \boldsymbol{\gamma}), \boldsymbol{\alpha}, \boldsymbol{\beta}, \boldsymbol{\gamma}) + \lambda\cdot{\rm Penalty}(\boldsymbol{\beta})\\
&{\rm s.t.} \; \mathbf{W}^*(\boldsymbol{\alpha}, \boldsymbol{\beta}, \boldsymbol{\gamma}) = {\rm argmin}_{\mathbf{W}}\mathcal{L}_{train}(\mathbf{W}, \boldsymbol{\alpha}, \boldsymbol{\beta}, \boldsymbol{\gamma})
\end{aligned}
\end{equation}
where $\boldsymbol{\alpha}$, $\boldsymbol{\beta}$, and $\boldsymbol{\gamma}$ are the architecture weights involved in the modality-specific search, feature selector, and searchable fusion, respectively. $\mathbf{W}$ denotes all other learnable parameters in the network, and $\mathcal{L}_{val}$ and $\mathcal{L}_{train}$ mean the validation loss and training loss.

Additionally, we add one more penalty term ${\rm Penalty}(\boldsymbol{\beta})$ to the architecture weights of the feature selector $\boldsymbol{\beta}$ in order to make the feature selector select diverse modalities at different step nodes. We achieve that by maximizing the cross entropy of any two different steps of $\boldsymbol{\beta}$:
\begin{equation} \label{eq:penalty}
\begin{aligned}
    &{\rm Penalty}(\boldsymbol{\beta}) = - \sum_{c_1 = 1}^{C}\sum_{c_2 = 1}^{C}\\
    &{\rm CrossEntropy}(\boldsymbol{\beta}[c_1][:4], \boldsymbol{\beta}[c_2][:4])
\end{aligned}
\end{equation}

Note that we only take the first four values of $\boldsymbol{\beta}$ at each step, which indicates the selection of input modalities $[\mathbf{z}_1, \cdots, \mathbf{z}_4]$. In this way, the penalty term will lead the feature selector to select different combinations of input modalities for fusion at different steps. However, the selection of the following computation nodes $[\mathbf{g}_1, \cdots, \mathbf{g}_{C}]$ will not be affected, which enables the model to search for more complex fusion strategies without any constraints.

\begin{algorithm}[t]
        \caption{Pruning Supernet}
        \label{algo:prune}
        \small
\begin{algorithmic}
\Input {A pretrained supernet $\mathcal{S}$, Set of mixed operations $\mathcal{E}$ from $\mathcal{S}$}
\Output {Set of selected operations $\{o_e^* \}_{e\in\mathcal{E}}$, Pruned supernet $\mathcal{S}^\prime$}
\While{True}
{
        \State randomly select a mixed operation $e \in \mathcal{E}$\;
        \For{all operation $o$ on $e$}
            \State evaluate the validation performance of $S$ when $o$ is removed ($\mathcal{S}_{\backslash o}$)\;
        \EndFor
        \State remove the worst operation for $e$: $o_e^\prime \gets \argmax_{o} {\rm Evaluate}(\mathcal{S}_{\backslash o})$\; 
        \State re-normalize architecture weights on $e$ for remaining operations\; 
        \State finetune the remaining supernet $\mathcal{S}^\prime$ for a few steps\; 
        \If{$\forall e\in \mathcal{E}, |\{o | o\in e\}| = 1$}
        \State break\;
        \EndIf
        }
        \EndWhile
\end{algorithmic}
\end{algorithm}

\subsubsection{Deriving the Optimal Architecture}\label{sec:prune}
After the supernet training, we propose a new way to derive the final architecture that gradually prunes operations from the supernet. We introduce the process of pruning in Algorithm \ref{algo:prune}. 
Specifically, we denote the trained supernet as $\mathcal{S}$ and the set of all mixed operations as $\mathcal{E}$. The goal is to prune the supernet until only one operation $o_e^*$ is left for each mixed operation $e\in\mathcal{E}$. 
In order to achieve that without too much performance drop from the supernet, we propose to gradually prune unimportant operations from the supernet, where the importance of each operation is determined by its contribution to the validation performance. 

For each iteration, we first sample a mixed operation from the supernet and then prune the worst operation from it if removing the operation brings the best validation performance. Then, we will fine-tune the remaining architecture until it converges again. The pruning will stop when all mixed operations from the supernet only have one optimal operation remaining. In this way, we can observe how the performance drops during the whole process, and select the relatively better architecture before it drops too much. In this way, we can get the final architecture with nearly the same performance as the trained supernet without training from scratch.



\section{Experiments}

\subsection{Experimental Setups}
\textbf{Data Processing}. \label{sec:data}
Following FIDDLE \cite{tang2020democratizing}, we extract data from the MIMIC-III dataset~\cite{johnson2016mimic} and specifically focus on the 17,710 patients (23,620 ICU visits) recorded from 2008 to 2012. 
We extract the structured features within 48/12 hours using FIDDLE. These features are randomly divided into the training, validation, and testing sets in a 7:1.5:1.5 ratio. The unstructured texts within 48/12 hours are obtained by consolidating the latest notes of each category into one document and aligning them with the other structured features. 
The data statistics can be found in Table~\ref{tab:data}.

\begin{table}[t]
     \centering
      \caption{Statistics of the four datasets.}
     \label{tab:data}
     \resizebox{0.5\textwidth}{!}{
     \begin{tabular}{@{}l|cccc@{}}
\toprule
Task             & ARF & Shock & Mortality & Diagnoses \\ \midrule
\# data samples &     9,393    &      12,289     &      7,440         &        7,440       \\
Dimension of $\mathbf{p}$        &    96     &    97     &           96    &    96           \\
Dimension of $\mathbf{E}$       &    680     &      739     &       1,001        &       1,001        \\
Dimension of $\mathbf{M}$        &     4,452    &       5,056    &       6,726        &      6,726         \\
Recorded hours $T$&     12    &      12     &   48            &     48          \\ \bottomrule
\end{tabular}}
    \vspace{-0.1in}
 \end{table}

\smallskip
\noindent \textbf{Prediction Tasks}.
We aim to address four prediction tasks: (1) Acute Respiratory Failure at 12 hours (\emph{ARF 12h}), (2) Shock at 12 hours (\emph{Shock 12h}), (3) Mortality at 48 hours (\emph{Mortality 48h}), and (4) Diagnoses at 48 hours (\emph{Diagnoses 48h}). The first three tasks involve \textbf{binary} classification, aiming to predict whether patients will experience these severe conditions during their ICU stays based on the initial 12/48 hours of data. The last task focuses on predicting diagnosis codes upon hospital discharge based on the initial 48 hours of data, constituting a \textbf{multi-label} classification problem. After extracting the top three digits of ICD-9 codes\footnote{\url{https://www.cdc.gov/nchs/icd/icd9.htm}}, we have 1,025 disease labels for predictions. 

\begin{table*}[!htb]
    \centering
    \caption{Performance comparison on four tasks. The second-best results are marked by \underline{underline}.}

    \label{tab:performance}
    \resizebox{0.9\textwidth}{!}{
    \begin{tabular}{l||cc|cc|cc||ccc}
\toprule
Tasks        & \multicolumn{2}{c|}{ARF 12h} & \multicolumn{2}{c|}{Shock 12h} & \multicolumn{2}{c||}{Mortality 48h} & \multicolumn{3}{c}{Diagnoses 48h}\\ \midrule
Metrics     & AUROC   & AUPR    & AUROC   & AUPR    & AUROC    & AUPR   & R@10 & R@20 & R@30   \\ \midrule
LSTM        &0.7377	& 0.3268			&0.7364 &	\underline{0.2940} 	&		0.8827	&0.5427	    &     0.1516	&0.1932	&0.2224\\
CNN         &     0.7348    &   0.3214      &  0.7356       &     0.2775       &      \underline{0.8876}    &    \underline{0.5479}        &    0.1670&	0.2501	&0.2990  \\ 
LstmBert    &         0.7310&	0.3086		&	\underline{0.7403}&	0.2852		&0.8755	&0.5362	&\underline{0.2299}	&0.3320	&0.3994    \\
BertLstm    &     \underline{0.7441}	&0.3200		&	0.7391&	0.2912		&	0.8811	&0.5409	&0.2170	&0.3298	&0.4086     \\
CnnBert     &         0.7201	&0.2925	&			0.7348&	0.2891&				0.8853&	0.5447	 &  0.2206	&0.3110	&0.3693       \\
BertCnn     &  0.7389	&\underline{0.3292}		&	0.7268&	0.2776	&	0.8777&	0.5450	  &  0.2250	&0.3346	&0.4072      \\
EncoderBert &   0.7350	&0.3117		&			0.7377	&0.2857		&		0.8800	&0.5296	 &  0.1985	&0.2459	&0.2825       \\
BertEncoder &    0.6893&	0.2633	&			0.7164	&0.2818	&				0.8663&	0.5183	    &  0.2198	&\underline{0.3366}	&\underline{0.4108}    \\
StarBert    &   0.7238	&0.3194		&			0.7110&	0.2869		&			0.8729	&0.4878 &  0.2265	&0.3305	&0.3996   \\
BertStar    &   0.6824&	0.2641&	0.7044	&0.2825	& 0.8644&	0.4724&  0.2215	&0.3173	&0.4026      \\  
MUFASA      &         0.7362	&0.3088			&	0.7295&	0.2735&			0.8812	&0.5436       &  0.0903   &0.1156  & 0.1296          \\ 
\rowcolor{blue!20}
\model        &   \textbf{0.7565}	&\textbf{0.3593}		&	\textbf{0.7463}&	\textbf{0.3039}	&		\textbf{0.8900}&	\textbf{0.5514}	     &  \textbf{0.3338} &\textbf{0.4780}	&\textbf{0.5658}       \\ \bottomrule
\end{tabular}
}
\vspace{-0.2in}
\end{table*}

\smallskip
\noindent \textbf{Baselines}.
We include both handcrafted and automated models as our baselines. \textbf{Handcrafted models}: we include vanilla LSTM~\cite{hochreiter1997long} and CNN~\cite{gu2018recent} models as baselines. Following the settings in \cite{yang2021leverage}, we combine LSTM, CNN, Transformer~\cite{vaswani2017attention}, and Star-Transformer \cite{guo-etal-2019-star} with ClinicalBERT using Multimodal Adaptation Gate (MAG)~\cite{rahman2020integrating}, resulting in \textbf{eight} baseline models as shown in Table \ref{tab:performance}. 
\textbf{Automated models}: we also apply MUFASA \cite{xu2021mufasa} to our prediction tasks as a NAS baseline for multi-modal EHR data. 

\smallskip
\noindent \textbf{Implementation}.
During the supernet training stage, we employ the Adam optimizer~\cite{kingma2014adam} with a learning rate of $1e-4$ for the network parameter $\mathbf{W}$ and $1e-5$ for the architecture parameters $\boldsymbol{\alpha}$, $\boldsymbol{\beta}$, and $\boldsymbol{\gamma}$. The loss function is cross entropy. The batch size for all tasks is set to 64, and the hidden dimension $d_e$ is set to 256. To balance the bi-level optimization, the penalty term weight $\lambda$ is set to 0.1. In the first stage, the number of layers $K$ is set to 2, while in the second stage, the number of step nodes $C$ is set to 3. The entire supernet training takes less than one hour on an NVIDIA A100 GPU, demonstrating the efficiency of the search algorithm.
During the pruning of the trained supernet, we utilize the same settings, except for the fine-tuning phase, where we use a learning rate of $2e-6$.

\smallskip
\noindent \textbf{Evaluation Metrics}.
Following~\cite{yang2021leverage}, for three binary classification tasks, we use \textbf{AUROC} (Area
Under the Receiver Operating Characteristic curve) and \textbf{AUPR} (Area Under the Precision-Recall curve) to evaluate the performance. For the diagnosis prediction task, we use \textbf{Top-$K$ Recall} (R@$K$) instead and separately set $K$ to be 10, 20, or 30. For all methods, we run the experiments for \textbf{five times} and report the mean values for a fair comparison.

\subsection{Performance Evaluation}

The results of the four tasks are presented in Table~\ref{tab:performance}, where it can be observed that our method achieves the best performance across all metrics compared to the baseline approaches. These results demonstrate the superiority of our proposed method over existing state-of-the-art baselines.

For the three binary classification tasks aiming to predict emergent conditions within a short time window, real-time time-series data, such as vital signs, play a crucial role in achieving high performance. This importance of real-time data is further confirmed by our ablation study in Section~\ref{sec:ablation}. However, existing approaches fail to adequately distinguish the significance of such modalities and treat all modalities equally in the input, leading to subpar performance.

In the case of the multi-label diagnosis prediction task, where the model is required to predict from a vast set of 1,025 disease groups, the task complexity is higher compared to the other tasks. Consequently, models employing advanced fusion techniques outperform those using simple fusion methods. Notably, MUFASA performs significantly worse than other baselines in this task. This can be attributed to the large label space, which makes it challenging for MUFASA to converge to a favorable local optimum.



\begin{table}[t]
    \centering
    \caption{Ablation study on input modalities}
    \label{tab:ab-input}

    \resizebox{0.48\textwidth}{!}{
    \begin{tabular}{l|lccc}
\toprule
\multicolumn{1}{l|}{Task}       & ARF  & Shock  & Mortality & Diagnoses\\ \midrule
\multicolumn{1}{l|}{Metrics}    & AUPR  & AUPR  & AUPR & R@10 \\ \midrule
\rowcolor{blue!20}
\multicolumn{1}{l|}{\model}   &    \textbf{0.3593}    &    \textbf{0.3039}   &  \textbf{0.5514 }  &  \textbf{0.3338} \\
 - Demographic &0.3376& 0.2822 & 0.5553& 0.3310\\
 - Continuous Events &0.2966& 0.2310& 0.4010& 0.3074\\
 - Discrete Events & 0.3551& 0.3027 &0.5336 & 0.3294\\
 - Clinical Notes & 0.3362& 0.2773& 0.5501& 0.3265\\

\bottomrule
\end{tabular}
    }
    \vspace{-0.1in}

\end{table}

\subsection{Ablation Study on Input Modalities}\label{sec:ablation}
In this ablation study, our objective is to assess the importance of each modality in the final prediction by systematically removing them from the model input alternatively. The experimental results are presented in Table \ref{tab:ab-input}. It can be observed that removing any modality leads to a drop in performance, highlighting the significance of considering multimodal EHR data as input for the model. Notably, the time series modality predominantly encompasses vital signs information, which is a strong indicator of emergency conditions occurring during ICU stays and aids in accurate diagnoses. Consequently, removing continuous event features from the input leads to the largest performance drop compared to the original framework.

\begin{table}[t]
    \centering
    \caption{Results on different optimization methods. }
    \label{tab:optimization}

    \resizebox{0.48\textwidth}{!}{
    \begin{tabular}{l|cccc}
\toprule
Tasks       & ARF  & Shock  & Mortality & Diagnoses\\ \midrule
Metrics    & AUPR  & AUPR  & AUPR & R@10 \\ \midrule
\rowcolor{red!20}
Supernet       &  \textbf{0.3565 }     &   \textbf{0.3053}    &  \textbf{0.5549}   &  0.3251 \\
Supernet$\circleddash$   &    0.3445    &   0.3027    &   0.5473  &  \textbf{0.3257 }\\ \hline\hline
\rowcolor{blue!20}
\model   &    \textbf{0.3593}    &    \textbf{0.3039}   &  \textbf{0.5514}   &  \textbf{0.3338} \\ 
\textsc{AutoFM}$\circleddash$  &       0.3410 &     0.3031  &   0.5471  & 0.3316 \\ \bottomrule
\end{tabular}
    }
    \vspace{-0.1in}
\end{table}

\subsection{Effect of Feature Selection Penalty}
We introduce a new loss term in the optimization process to guide the selection of optimal modalities in the fusion stage. To assess the effectiveness of this designed penalty term (Eq.\eqref{eq:penalty}), we conducted a study, and the results are presented in Table~\ref{tab:optimization}.

``Supernet'' refers to the learned supernet using our proposed \model with Eq.~\eqref{eq:bilevel}, while ``Supernet$\circleddash$'' indicates the supernet without the penalty term defined in Eq.~\eqref{eq:penalty}.
We observe that the inclusion of the penalty term improves the performance of the supernet across all three binary classification tasks. For the diagnoses prediction task, comparable performance is still achieved even without the penalty term. These results confirm the term's ability to ensure the diversity of feature selection in the multi-modal fusion process. 

We also compare the derived model from Supernet$\circleddash$, denoted as ``\textsc{AutoFM}$\circleddash$'', which still employs the proposed pruning-based architecture discretization outlined in Algorithm~\ref{algo:prune}.
Notably, without the designed penalty, the derived model \textsc{AutoFM}$\circleddash$ exhibits inferior performance compared to our proposed \model. This comparison further demonstrates the necessity and effectiveness of the penalty term in the optimization process.

\begin{table}[t]
    \centering
    \caption{Results on different discretization methods. }
    \label{tab:pruning_results}

    \resizebox{0.48\textwidth}{!}{
    \begin{tabular}{l|cccc}
\toprule
Tasks       & ARF  & Shock  & Mortality & Diagnoses\\ \midrule
Metrics    & AUPR  & AUPR  & AUPR & R@10 \\ \midrule
DARTS &    0.3124    &    0.2891   &  \textbf{0.5536}   &  0.3332 \\
DARTS-PT      &  0.3433      &    0.2986   &   0.5444  & 0.3268  \\
\rowcolor{blue!20}
\model   &    \textbf{0.3593}    &    \textbf{0.3039}   &  0.5514   &  \textbf{0.3338} \\ \bottomrule
\end{tabular}
    }
    \vspace{-0.1in}
\end{table}

\subsection{Pruning-based Architecture Selection}
In Section~\ref{sec:prune}, we present a pruning-based discretization method to derive the optimal architecture. To evaluate the effectiveness of this approach, we derive two models from the same supernet trained using Eq.~\eqref{eq:bilevel}, employing the approaches used in DARTS~\cite{liu2018darts} and DARTS-PT~\cite{wang2021rethinking}. The results are shown in Table~\ref{tab:pruning_results}. 
The performance of the derived models using our pruning-based approach is significantly better compared to the models derived using DARTS and DARTS-PT. This indicates that our method effectively captures the optimal architecture from the supernet and achieves superior performance. These results provide strong evidence supporting the effectiveness of our proposed discretization method. 


\begin{figure}[t]
    \centering
    \resizebox{0.5\textwidth}{!}{
    \includegraphics{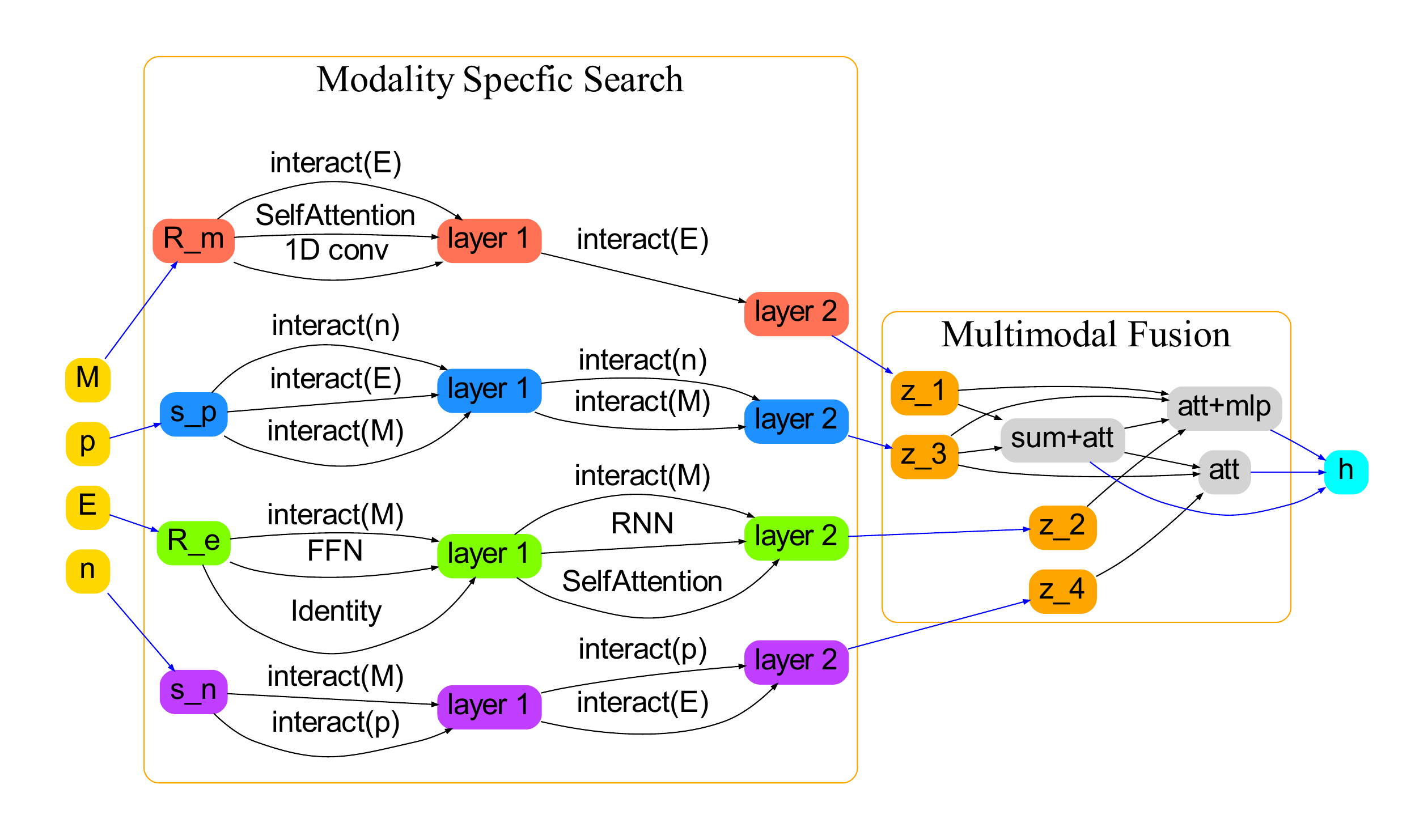}}
    \vspace{-0.2in}
    \caption{Searched architecture. The blue arrows represent fixed operations, while the other black arrows are all searched operations. The $\rm interact(\cdot)$ means the interaction operation with the corresponding feature. For the steps nodes $[\mathbf{g}_1, \mathbf{g}_2, \mathbf{g}_3]$, we omit the notations in the figure and fill the node with the selected fusion operations like (sum+att). }
    \label{fig:arch}
\vspace{-0.2in}
\end{figure}

\subsection{Architecture Study}
Figure~\ref{fig:arch} provides an example of the searched architecture for the ARF task. It demonstrates the effectiveness of our approach in capturing the interactions between different modalities during the modality-specific search stage. 
Furthermore, in the second stage of multi-modal fusion, our method is able to explore different combinations of input modalities $[\mathbf{z}_1, \cdots, \mathbf{z}_4]$ at different steps. This is achieved through the penalty imposed by the additional loss term, which encourages diversity in the feature selection process. As a result, our fusion module can effectively determine the optimal fusion strategy, leading to improved performance in capturing the underlying patterns and relationships within the multi-modal EHR data.  

\section{Conclusion}
In this paper, we introduced \model, a novel Neural Architecture Search (NAS) framework designed for automatically fusing multi-modal EHR data. Our experimental results demonstrated the superior performance of our method compared to existing state-of-the-art baselines across various tasks. 

\section*{Acknowledgement}
This work is partially supported by the National Science Foundation (NSF) under Grant No. 1951729, 1953813, 2119331, 2333790, 2212323, and 2238275, and the National Institutes of Health (NIH) under Grant No. R01AG077016.

\bibliographystyle{plain}
\bibliography{ref}

\begin{thebibliography}{10}

\bibitem{choi2016doctor}
Edward Choi, Mohammad~Taha Bahadori, Andy Schuetz, Walter~F Stewart, and Jimeng Sun.
\newblock Doctor ai: Predicting clinical events via recurrent neural networks.
\newblock In {\em Machine learning for healthcare conference}, pages 301--318. PMLR, 2016.

\bibitem{choi2016retain}
Edward Choi, Mohammad~Taha Bahadori, Jimeng Sun, Joshua Kulas, Andy Schuetz, and Walter Stewart.
\newblock Retain: An interpretable predictive model for healthcare using reverse time attention mechanism.
\newblock In {\em Advances in Neural Information Processing Systems}, pages 3504--3512, 2016.

\bibitem{chung2014empirical}
Junyoung Chung, Caglar Gulcehre, KyungHyun Cho, and Yoshua Bengio.
\newblock Empirical evaluation of gated recurrent neural networks on sequence modeling.
\newblock {\em arXiv preprint arXiv:1412.3555}, 2014.

\bibitem{cui2022automed}
Suhan Cui, Jiaqi Wang, Xinning Gui, Ting Wang, and Fenglong Ma.
\newblock Automed: Automated medical risk predictive modeling on electronic health records.
\newblock In {\em 2022 IEEE International Conference on Bioinformatics and Biomedicine (BIBM)}, pages 948--953. IEEE, 2022.

\bibitem{elsken2019neural}
Thomas Elsken, Jan~Hendrik Metzen, and Frank Hutter.
\newblock Neural architecture search: A survey.
\newblock {\em The Journal of Machine Learning Research}, 20(1):1997--2017, 2019.

\bibitem{feng2019dcmn}
Yujuan Feng, Zhenxing Xu, Lin Gan, Ning Chen, Bin Yu, Ting Chen, and Fei Wang.
\newblock Dcmn: Double core memory network for patient outcome prediction with multimodal data.
\newblock In {\em 2019 IEEE International Conference on Data Mining (ICDM)}, pages 200--209. IEEE, 2019.

\bibitem{gu2018recent}
Jiuxiang Gu, Zhenhua Wang, Jason Kuen, Lianyang Ma, Amir Shahroudy, Bing Shuai, Ting Liu, Xingxing Wang, Gang Wang, Jianfei Cai, et~al.
\newblock Recent advances in convolutional neural networks.
\newblock {\em Pattern recognition}, 77:354--377, 2018.

\bibitem{guo-etal-2019-star}
Qipeng Guo, Xipeng Qiu, Pengfei Liu, Yunfan Shao, Xiangyang Xue, and Zheng Zhang.
\newblock Star-transformer.
\newblock In {\em Proceedings of the 2019 Conference of the North {A}merican Chapter of the Association for Computational Linguistics: Human Language Technologies, Volume 1 (Long and Short Papers)}, pages 1315--1325, Minneapolis, Minnesota, June 2019. Association for Computational Linguistics.

\bibitem{hochreiter1997long}
Sepp Hochreiter and J{\"u}rgen Schmidhuber.
\newblock Long short-term memory.
\newblock {\em Neural computation}, 9(8):1735--1780, 1997.

\bibitem{huang2019clinicalbert}
Kexin Huang, Jaan Altosaar, and Rajesh Ranganath.
\newblock Clinicalbert: Modeling clinical notes and predicting hospital readmission.
\newblock {\em arXiv preprint arXiv:1904.05342}, 2019.

\bibitem{johnson2016mimic}
Alistair~EW Johnson, Tom~J Pollard, Lu~Shen, Li-wei~H Lehman, Mengling Feng, Mohammad Ghassemi, Benjamin Moody, Peter Szolovits, Leo Anthony~Celi, and Roger~G Mark.
\newblock Mimic-iii, a freely accessible critical care database.
\newblock {\em Scientific data}, 3(1):1--9, 2016.

\bibitem{kandasamy2018neural}
Kirthevasan Kandasamy, Willie Neiswanger, Jeff Schneider, Barnabas Poczos, and Eric~P Xing.
\newblock Neural architecture search with bayesian optimisation and optimal transport.
\newblock {\em Advances in neural information processing systems}, 31, 2018.

\bibitem{kingma2014adam}
Diederik~P Kingma and Jimmy Ba.
\newblock Adam: A method for stochastic optimization.
\newblock {\em arXiv preprint arXiv:1412.6980}, 2014.

\bibitem{liu2018progressive}
Chenxi Liu, Barret Zoph, Maxim Neumann, Jonathon Shlens, Wei Hua, Li-Jia Li, Li~Fei-Fei, Alan Yuille, Jonathan Huang, and Kevin Murphy.
\newblock Progressive neural architecture search.
\newblock In {\em Proceedings of the European conference on computer vision (ECCV)}, pages 19--34, 2018.

\bibitem{liu2018darts}
Hanxiao Liu, Karen Simonyan, and Yiming Yang.
\newblock Darts: Differentiable architecture search.
\newblock In {\em International Conference on Learning Representations}, 2019.

\bibitem{ma2017dipole}
Fenglong Ma, Radha Chitta, Jing Zhou, Quanzeng You, Tong Sun, and Jing Gao.
\newblock Dipole: Diagnosis prediction in healthcare via attention-based bidirectional recurrent neural networks.
\newblock In {\em Proceedings of the 23rd ACM SIGKDD international conference on knowledge discovery and data mining}, pages 1903--1911, 2017.

\bibitem{ma2018risk}
Fenglong Ma, Jing Gao, Qiuling Suo, Quanzeng You, Jing Zhou, and Aidong Zhang.
\newblock Risk prediction on electronic health records with prior medical knowledge.
\newblock In {\em Proceedings of the 24th ACM SIGKDD International Conference on Knowledge Discovery \& Data Mining}, pages 1910--1919, 2018.

\bibitem{pham2018efficient}
Hieu Pham, Melody Guan, Barret Zoph, Quoc Le, and Jeff Dean.
\newblock Efficient neural architecture search via parameters sharing.
\newblock In {\em International conference on machine learning}, pages 4095--4104. PMLR, 2018.

\bibitem{qiao2019mnn}
Zhi Qiao, Xian Wu, Shen Ge, and Wei Fan.
\newblock Mnn: multimodal attentional neural networks for diagnosis prediction.
\newblock {\em Extraction}, 1:A1, 2019.

\bibitem{rahman2020integrating}
Wasifur Rahman, Md~Kamrul Hasan, Sangwu Lee, Amir Zadeh, Chengfeng Mao, Louis-Philippe Morency, and Ehsan Hoque.
\newblock Integrating multimodal information in large pretrained transformers.
\newblock In {\em Proceedings of the conference. Association for Computational Linguistics. Meeting}, volume 2020, page 2359. NIH Public Access, 2020.

\bibitem{real2019regularized}
Esteban Real, Alok Aggarwal, Yanping Huang, and Quoc~V Le.
\newblock Regularized evolution for image classifier architecture search.
\newblock In {\em Proceedings of the aaai conference on artificial intelligence}, volume~33, pages 4780--4789, 2019.

\bibitem{shang2019gamenet}
Junyuan Shang, Cao Xiao, Tengfei Ma, Hongyan Li, and Jimeng Sun.
\newblock Gamenet: Graph augmented memory networks for recommending medication combination.
\newblock In {\em proceedings of the AAAI Conference on Artificial Intelligence}, volume~33, pages 1126--1133, 2019.

\bibitem{tang2020democratizing}
Shengpu Tang, Parmida Davarmanesh, Yanmeng Song, Danai Koutra, Michael~W Sjoding, and Jenna Wiens.
\newblock Democratizing ehr analyses with fiddle: a flexible data-driven preprocessing pipeline for structured clinical data.
\newblock {\em Journal of the American Medical Informatics Association}, 27(12):1921--1934, 2020.

\bibitem{vaswani2017attention}
Ashish Vaswani, Noam Shazeer, Niki Parmar, Jakob Uszkoreit, Llion Jones, Aidan~N Gomez, {\L}ukasz Kaiser, and Illia Polosukhin.
\newblock Attention is all you need.
\newblock In {\em Advances in neural information processing systems}, pages 5998--6008, 2017.

\bibitem{wang2021rethinking}
Ruochen Wang, Minhao Cheng, Xiangning Chen, Xiaocheng Tang, and Cho-Jui Hsieh.
\newblock Rethinking architecture selection in differentiable nas.
\newblock {\em arXiv preprint arXiv:2108.04392}, 2021.

\bibitem{xu2018raim}
Yanbo Xu, Siddharth Biswal, Shriprasad~R Deshpande, Kevin~O Maher, and Jimeng Sun.
\newblock Raim: Recurrent attentive and intensive model of multimodal patient monitoring data.
\newblock In {\em Proceedings of the 24th ACM SIGKDD international conference on Knowledge Discovery \& Data Mining}, pages 2565--2573, 2018.

\bibitem{xu2021mufasa}
Zhen Xu, David~R So, and Andrew~M Dai.
\newblock Mufasa: Multimodal fusion architecture search for electronic health records.
\newblock In {\em Proceedings of the AAAI Conference on Artificial Intelligence}, volume~35, pages 10532--10540, 2021.

\bibitem{yang2021leverage}
Bo~Yang and Lijun Wu.
\newblock How to leverage multimodal ehr data for better medical predictions?
\newblock {\em arXiv preprint arXiv:2110.15763}, 2021.

\bibitem{zela2019understanding}
Arber Zela, Thomas Elsken, Tonmoy Saikia, Yassine Marrakchi, Thomas Brox, and Frank Hutter.
\newblock Understanding and robustifying differentiable architecture search.
\newblock {\em arXiv preprint arXiv:1909.09656}, 2019.

\bibitem{zhang2022improving}
Xinlu Zhang, Shiyang Li, Zhiyu Chen, Xifeng Yan, and Linda Petzold.
\newblock Improving medical predictions by irregular multimodal electronic health records modeling.
\newblock {\em arXiv preprint arXiv:2210.12156}, 2022.

\bibitem{zoph2016neural}
Barret Zoph and Quoc~V Le.
\newblock Neural architecture search with reinforcement learning.
\newblock {\em arXiv preprint arXiv:1611.01578}, 2016.

\end{thebibliography}
\end{document}